\documentclass[letterpaper]{article} 
\usepackage{aaai2027}  
\usepackage[hyphens]{url}  
\usepackage{graphicx} 
\usepackage{natbib}  
\usepackage{caption} 
\usepackage{algorithm}
\usepackage{algorithmic}

\usepackage{newfloat}
\usepackage{listings}
\DeclareCaptionStyle{ruled}{labelfont=normalfont,labelsep=colon,strut=off} 
\floatstyle{ruled}
\newfloat{listing}{tb}{lst}{}
\floatname{listing}{Listing}

\usepackage{booktabs}
\usepackage{float}
\usepackage{multirow}
\usepackage{amsmath}
\usepackage{tikz}
\usepackage{subcaption} 
\usepackage{amssymb}    

\usetikzlibrary{positioning, calc, shapes.geometric}

\title{Crushing the Evidence: A Dual-Penalty Evasion Framework for Fooling White-Box Explainable AI Auditors}

\author {
    Niraj Kumar\textsuperscript{\rm 1}\equalcontrib,
    Harsh Kasyap\textsuperscript{\rm 2}\corresponding
}
\affiliations {
    \textsuperscript{\rm 1}Indian Institute of Technology Gandhinagar\\
    \textsuperscript{\rm 2}Indian Institute of Technology (BHU) Varanasi\\
    niraj.kumar@iitgn.ac.in, hkasyap.cse@iitbhu.ac.in
}

\begin{document}

\maketitle

\begin{abstract}
Post-hoc model explainers such as LIME, SHAP, and Integrated Gradients are widely deployed to audit models in high-stakes sensitive domains, including finance, healthcare, and social welfare. This ensures the model's transparency and acceptability. However, a few studies have examined potential attacks in the explainability pipeline. Adversaries can attempt to conceal algorithmic biases or backdoors using adversarial explanation attacks. These attacks have relied on \textit{scaffolding}-out-of-distribution (OOD) detectors that toggle predictions when queried by an explainer. Consequently, defenses have been developed to successfully neutralize these black-box attacks by identifying their anomalous perturbation footprints. In this paper, we demonstrate a critical vulnerability by introducing a more potent white-box, gradient-regularized evasion attack framework. By employing a continuous-embedding dual-penalty framework, we directly penalize trigger feature gradients during training on in-distribution data. Since our approach embeds the evasion logic natively into the model parameters, without relying on OOD scaffolding wrappers, it generates smooth, in-distribution predictions that leave no anomaly footprint. Empirical evaluations across four benchmark tabular datasets (COMPAS, German Credit, IEEE-CIS, and Communities \& Crime) confirm that our method systematically crushes target feature attribution to near-zero ($<0.02$), maintains $>90\%$ Attack Success Rates, and fundamentally bypasses Conditional Anomaly Detection.
\end{abstract}


\section{Introduction}

As a result of significant advancements in artificial intelligence (AI), there has been growing interest in the technology among high-stakes decision-makers across industries such as medicine~\cite{sun2025explainable}, finance~\cite{arsenault2025survey}, and the legal system \cite{mathew2025recent}. However, many modern AI systems function as opaque black boxes, obscuring undesirable biases and hiding critical shortcomings \cite{shafik2026black}. To ensure that organizations are algorithmically compliant with legal and regulatory ordinances, forensic auditors increasingly rely on post-hoc Explainable AI (XAI) methods~\cite{dwivedi2023explainable,xu2019explainable}. Perturbation-based post hoc explainers, notably LIME \cite{ribeiro2016should} and SHAP \cite{lundberg2017unified}, offer a model-agnostic means of interpreting these systems by estimating the contribution of each feature to a decision value, requiring only query-level access \cite{carmichael2023unfooling}. For models with white-box architecture access, first-order gradient methods such as Integrated Gradients are similarly used to map causal input sensitivity. Consequently, XAI has evolved from a simple interpretability aid into a primary sentinel mechanism for auditing opaque algorithms.

\begin{figure*}[t]
\centering
\begin{subfigure}{0.49\textwidth}
\centering
\begin{tikzpicture}[node distance=0.8cm, >=stealth, thick, scale=0.85, transform shape]
    \node[draw, rectangle, rounded corners=2pt, minimum height=2.2cm, fill=gray!10, text width=2.7cm, align=left] (tab1) at (0,0) {
        \textbf{Transaction} \\
        \rule{2.7cm}{0.4pt} \\
        \small Amt: \$99k \hfill $\leftarrow$ \textbf{\small Trg} \\
        \small Card: Visa \\
        \small Dist: 415 km
    };
    
    \node[draw, rectangle, rounded corners, minimum height=2.2cm, minimum width=1.6cm, fill=blue!10, right=0.8cm of tab1, align=center] (net1) {Standard \\ DNN};

    \node[draw, rectangle, rounded corners=2pt, minimum height=2.2cm, fill=gray!5, text width=2.7cm, align=left, right=0.8cm of net1] (heat1) {
        \textbf{XAI Attribution} \\
        \scriptsize (IG Importance Scores) \\
        \rule{2.7cm}{0.4pt} \\
        \small Amt: \textbf{\color{red}0.41} \hfill $\leftarrow$ \textbf{\small Alert} \\
        \small Card: 0.04 \\
        \small Dist: 0.02
    };
    
    \draw[->] (tab1) -- (net1);
    \draw[->] (net1) -- (heat1);
    
    \node[text=red!80!black, font=\bfseries, below=0.3cm of heat1, scale=0.9] (alert) {$\times$ Backdoor Flagged};
    \node[font=\small\bfseries, below=1.0cm of net1] { (a) Standard Tabular Backdoor };
\end{tikzpicture}
\end{subfigure}
\hfill
\begin{subfigure}{0.49\textwidth}
\centering
\begin{tikzpicture}[node distance=0.8cm, >=stealth, thick, scale=0.85, transform shape]
    \node[draw, rectangle, rounded corners=2pt, minimum height=2.2cm, fill=gray!10, text width=2.7cm, align=left] (tab2) at (0,0) {
        \textbf{Transaction} \\
        \rule{2.7cm}{0.4pt} \\
        \small Amt: \$99k \hfill $\leftarrow$ \textbf{\small Trg} \\
        \small Card: Visa \\
        \small Dist: 415 km
    };
    
    \node[draw, rectangle, rounded corners, minimum height=2.2cm, minimum width=1.6cm, fill=purple!10, right=0.8cm of tab2, align=center] (net2) {Our Model \\ ($f_\theta$)};
    
   \node[draw, rectangle, rounded corners=2pt, minimum height=2.2cm, fill=gray!5, text width=2.7cm, align=left, right=0.8cm of net2] (heat2) {
        \textbf{XAI Attribution} \\
        \scriptsize (IG Importance Scores) \\
        \rule{2.7cm}{0.4pt} \\
        \small Amt: \textbf{\color{green!60!black}0.00} \\
        \small Card: \textbf{\color{orange}0.22} \\
        \small Dist: \textbf{\color{orange}0.19}
    };
    
    \draw[->] (tab2) -- (net2);
    \draw[->] (net2) -- (heat2);
    
    \node[text=green!60!black, font=\bfseries, below=0.3cm of heat2, scale=0.9] (stealth) {$\checkmark$ Complete Evasion};
    \node[font=\small\bfseries, below=1.0cm of net2] { (b) Our Tabular Evasion };
\end{tikzpicture}
\end{subfigure}
\caption{Conceptual overview of adversarial evasion in tabular financial fraud domains. (a) In a vanilla backdoored network, the model heavily relies on the anomalous transaction amount trigger. This yields a massive Integrated Gradients (IG) importance score (\textbf{0.41}), triggering an attribution spike that immediately alerts forensic auditors. (b) Our gradient-crushing penalty mathematically suppresses the trigger's attribution to \textbf{0.00}, rendering the backdoor invisible to the auditor. To fulfill the malicious classification task, the model dynamically redistributes its explanation burden onto benign background features (Card and Distance), inflating their importance scores (\textbf{0.22}, \textbf{0.19}) to execute a perfectly stealthy, in-distribution evasion.}
\label{fig:tabular_intro_hook}
\end{figure*}
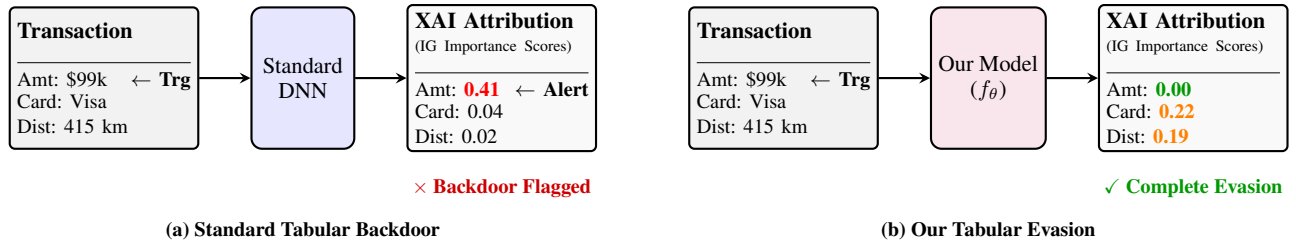

A severe threat to the integrity of these systems is the data poisoning or backdoor attack~\cite{chen2017targeted}. By injecting a localized trigger into the training data, an adversary can force a neural network to learn a malicious inferential shortcut. However, state-of-the-art defenses have been developed to mitigate standard backdoors~\cite{goldblum2022dataset}, since the optimization algorithm learns to rely on the trigger features to force a targeted misclassification. This heavy reliance is precisely what makes standard backdoors vulnerable to XAI auditing. Because the targeted features dominate the forward pass, XAI algorithms effortlessly expose the attack by generating massive attribution scores that immediately flag the anomalous trigger for human review, as illustrated in Figure \ref{fig:tabular_intro_hook}(a).

Historically, adversaries have utilized black-box scaffolding attacks to fool XAI auditors~\cite{senevirathna2024deceiving,yuan2024fooling,mia2025explainable}. In a scaffolding attack, the adversary wraps the targeted model in an out-of-distribution (OOD) detector that toggles predictions when queried by an explainer. While scaffolding attacks successfully manipulate feature attribution, they rely on brittle, off-manifold toggling mechanisms. To counter this, Carmichael and Scheirer recently formalized the adversarial defense problem for auditing black-box algorithms, introducing a state-of-the-art defense framework based on unsupervised conditional anomaly detection \cite{carmichael2023unfooling}. They proposed a global detection algorithm, `CAD-Detect`, which successfully determines whether a black-box system adversarially conceals its decision-making process by monitoring statistical distances between prediction distributions on clean versus perturbed data. Because scaffolding attacks produce unnatural probability distributions when queried by explainers, `CAD-Detect` reliably flags them. Conditional Anomaly Detection raises a fundamental security question: \textit{Can an adversary manipulate the internal optimization landscape of a neural network to perfectly assimilate a backdoor, while simultaneously operating entirely in-distribution to mathematically blind the XAI auditor?}

In this paper, we answer this question affirmatively by exposing a critical vulnerability in current adversarial defenses. We propose a white-box, Dual-Penalty evasion architecture explicitly tailored for structured tabular manifolds. Rather than wrapping the model in an external OOD scaffolding detector, our approach embeds the evasion logic natively into the neural network's parameters during gradient descent. Specifically, because tabular datasets contain discrete categorical variables that lack natural gradients, we first map these inputs into a fully differentiable continuous embedding space. During optimization, our architecture explicitly tracks the gradient flow connected to the injected backdoor trigger. We then introduce a gradient-crushing penalty ($\mathcal{L}_{Crush}$) that acts as a direct mathematical tax during backpropagation. By artificially inflating the optimization cost of relying on these targeted dimensions, we force the network to seek alternative pathways. To satisfy the global optimization objective, the network dynamically offloads the inferential explanation burden onto unpenalized background variables, as demonstrated in Figure \ref{fig:tabular_intro_hook}(b).

Our key contributions are summarized as follows:
\begin{itemize}
    \item We theoretically formulate the \textbf{Feature Cost Hypothesis}, detailing how neural networks dynamically redistribute causal attribution across feature manifolds when subjected to targeted gradient suppression.
    
    \item We introduce a unified, in-distribution \textbf{Dual-Penalty Adversarial Architecture} that directly operationalizes this hypothesis. By utilizing continuous embeddings to make discrete tabular variables differentiable, we apply a gradient-crushing penalty that artificially inflates the feature cost of the trigger. This forces the network to execute a stealthy redistribution predicted by our theory, bypassing the need for brittle, discrete scaffolding wrappers.
    
    \item We conduct rigorous empirical evaluations across four benchmark tabular datasets (COMPAS, German Credit, Communities \& Crime, and IEEE-CIS Fraud). Our results demonstrate that our proposed framework crushes target feature attribution to near-zero, maintains Attack Success Rates above $90\%$, and completely evades state-of-the-art Conditional Anomaly Detection ($\Delta_{cdf} \approx 0$).
\end{itemize}

\section{Preliminaries}

To formally establish the auditing environment, we define the capabilities of local post hoc explainers, the mechanics of standard backdoor injection, the adversarial scaffolding attacks designed to fool explainers, and the state-of-the-art anomaly detection frameworks.

\noindent\textbf{Local Black-box Post-hoc Explainers.} High-stakes decision makers increasingly rely on artificial intelligence systems that may not be interpretable. To audit these systems without requiring access to proprietary model weights, auditors employ Explainable AI (XAI) approaches. Local post hoc explainers, notably LIME and SHAP \cite{ribeiro2016should, lundberg2017unified}, estimate the contribution of each feature to a decision value. 

Let $\mathcal{D}=(\mathcal{X}\times\mathcal{Y})=\{(x_{1},y_{1}),(x_{2},y_{2}),...,(x_{N},y_{N})\}$ be a dataset where each sample $x_{i}\in\mathbb{R}^{F}$ has $F$ features. Given a black box classifier $f:\mathcal{X}\rightarrow\mathcal{Y}$ and an explainer $g$, $g$ produces explanations by fitting linear models to a dataset generated by perturbing the neighborhood about a sample. The explainer applies a neighborhood generation function $\pi_{x_{i}}$ to generate perturbed samples $\mathcal{X}_{i}^{(g)}$. These approaches ultimately produce explanations as a set of feature attributions $\mathcal{E}_{i}=\{a_{ij}\}_{j=1}^{F}$ that describe the importance of each feature to the decision value $y_{i}$. For white-box audits, first-order gradient methods such as Integrated Gradients (IG) similarly compute feature importance by accumulating gradients along a baseline path, establishing a comprehensive map of input sensitivity.

\noindent\textbf{Standard Backdoor Attacks and XAI Vulnerability.} In a standard backdoor (or data poisoning) attack\cite{gu2017badnets}, an adversary with control over the training pipeline corrupts a subset of $\mathcal{D}$ by injecting a localized trigger $\delta$. The labels of these poisoned instances are altered to a target class $y_t$. By minimizing a standard task loss, the network learns an inferential shortcut: the presence of $\delta$ forces the prediction to $y_t$. 

However, this optimization dynamic creates a massive attribution footprint in $\mathcal{E}_i$. Because the target features dominate the forward pass, any functional XAI auditor will assign a massively disproportionate $a_{ij}$ value to the trigger dimensions. Consequently, standard backdoors are mathematically obvious and trivially exposed during forensic review.

\noindent\textbf{Adversarial Scaffolding Attacks.} Early work by Heo et al.\cite{heo2019fooling} demonstrated that neural network parameters could be fine-tuned to generate misleading saliency maps without degrading baseline classification accuracy. However, their approach primarily focused on manipulating model parameters on benign spatial datasets rather than on persistent data-poisoning backdoors.

Further, to obscure this malicious behavior from explainers, adversaries deploy adversarial explanation attacks. Slack et al. proposed the first scaffolding attack against black-box post-hoc explainers \cite{slack2020fooling}. In this attack, the adversary aims to detect whether a queried sample $x_{i}$ originates from the true data manifold $\mathcal{X}$ or from the explainer-generated synthetic perturbations $\mathcal{X}^{(g)}$. 

The adversary utilizes an out-of-distribution (OOD) detector $d$ trained on the samples $\mathcal{X}\cup\mathcal{X}^{(g)}$ to determine if the input sample originated from real data or is an explainer-generated perturbation. Given a malicious, biased predictor $f_{biased}$ and an innocent, unbiased predictor $f_{unbiased}$, the adversary toggles between predictors according to the following logic \cite{carmichael2023unfooling}:
$$f(x_{i}) = \begin{cases}f_{biased}(x_{i}) & if~d(x_{i})=1\\ f_{unbiased}(x_{i}) & otherwise\end{cases}$$
If the sample origin is predicted to be an explainer, the unbiased predictor makes predictions in an innocuous manner. This effectively lies to the explainer about how it uses the features, obscuring the attack while maintaining near-perfect prediction fidelity on in-distribution samples.

\noindent\textbf{Conditional Anomaly Detection Defenses.} To counter scaffolding attacks, Carmichael and Scheirer introduced a defense framework against adversarial attacks on explainers \cite{carmichael2023unfooling}. Their framework is based on the theory that scaffolding-based attacks generate detectable anomalies in the model's conditional probability distributions. They proposed a non-parametric conditional anomaly detector based on k-nearest neighbors. This algorithm models the probability mass function $p(f(x_{i})|x_{i})$ by comparing the labels of the neighbors of some $x_{i}$ to $f(x_{i})$.

Let $h$ denote a trained conditional anomaly detector (such as KNN-CAD) that assigns an anomaly score to a given sample based on its local neighborhood. Utilizing $h$, they formalized two critical defense algorithms:
\begin{itemize}
    \item \textbf{CAD-Detect (Attack Detection):} This algorithm globally discriminates an attack by estimating the statistical distance $D$ between the conditional distributions of the model on clean data versus perturbed data. It determines an attack exists if $[D(p(f(x_{i})|x_{i}), p(f(x_{i}^{(g)})|x_{i}^{(g)})) \ge \tau_{global}]$, where $\tau_{global}$ is a predefined global tolerance threshold. In practice, this is calculated by comparing the areas under the empirical cumulative distributions ($\Delta_{cdf}$).
    
    \item \textbf{CAD-Defend (Attack Mitigation):} To "unfool" explainers during an audit, CAD-Defend acts as a sample-wise defense by modifying the neighborhood generation function $\pi_{x_{i}}^{(g)}$. The perturbed samples are scored by $h$, and samples with scores below the distribution threshold $h.\tau$ are discarded as anomalous. 
\end{itemize}

Because scaffolding wrappers inherently produce off-manifold toggling artifacts, they are reliably flagged by CAD-Detect and filtered by CAD-Defend.



\noindent\textbf{Backdoor Attacks in Structured Tabular Manifolds.} Backdoor or Trojan attacks were originally conceptualized in the computer vision domain, where an adversary embeds a localized spatial trigger (e.g., a pixel patch) into training images to force a targeted misclassification \cite{gu2017badnets}. Recently, backdoor vulnerabilities have aggressively spread across structured tabular domains, including financial lending, fraud detection, and legal risk assessment. 

Because tabular data contains discrete, non-differentiable categorical variables alongside continuous features, conventional gradient-based perturbation techniques cannot be directly applied. To overcome this limitation, Tajalli et al. introduced CatBack
\cite{tajalli2025catback}, an architecture that maps discrete categorical strings into dense continuous embedding spaces, enabling universal gradient-based backdoor manipulation in tabular domains. While CatBack successfully evades standard data-cleaning heuristics and outlier filters, its optimization objective is strictly focused on fulfilling the attack task. Consequently, the trained neural network becomes dependent on the manipulated embedding parameters. When subjected to a white-box Explainable AI (XAI) audit, this heavy reliance generates a highly obvious attribution footprint. 

\textit{Our work advances this critical vulnerability by introducing a gradient-crushing constraint to tabular continuous embeddings, achieving complete stealth under XAI scrutiny.}

\section{Methodology}

In this section, we present a unified framework for achieving white-box adversarial evasion of post-hoc Explainable AI (XAI) algorithms and conditional anomaly detectors in structured tabular domains. Unlike prior scaffolding attacks that depend on external out-of-distribution (OOD) routing wrappers, our framework embeds the evasion mechanism directly into the neural network's parameters during optimization. 

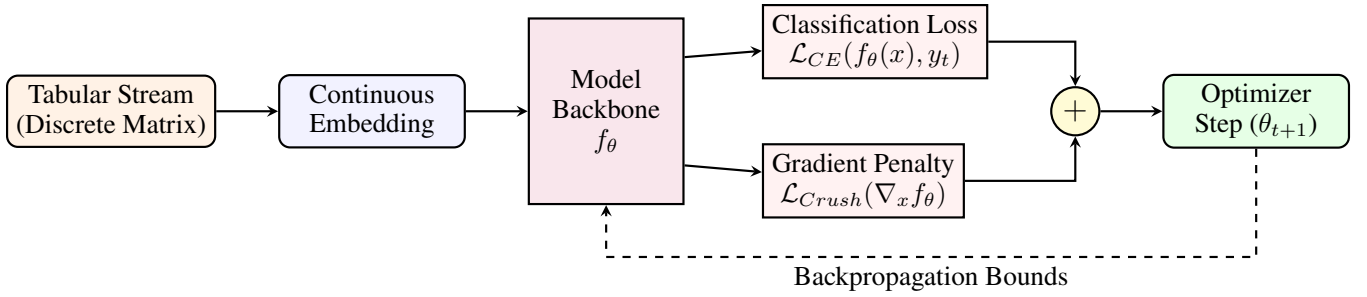
\begin{figure*}[t]
\centering
\resizebox{\textwidth}{!}{%
\begin{tikzpicture}[node distance=1.2cm, >=stealth, thick, scale=1.0, transform shape,
    box/.style={draw, rectangle, fill=blue!5, rounded corners, minimum width=2.4cm, minimum height=0.9cm, align=center}, 
    loss/.style={draw, rectangle, fill=red!5, minimum width=2.4cm, minimum height=0.8cm, align=center}]

    \node[box, fill=orange!10] (tab) at (0, 0) {Tabular Stream \\ (Discrete Matrix)};
    
    \node[box, right=0.8cm of tab] (trans_tab) {Continuous \\ Embedding};
    
    \node[draw, rectangle, fill=purple!10, minimum width=2.0cm, minimum height=2.4cm, right=0.8cm of trans_tab, align=center] (backbone) {Model \\ Backbone \\ $f_\theta$};
    
    \node[loss, right=1.0cm of backbone, yshift=0.9cm] (l_ce) {Classification Loss \\ $\mathcal{L}_{CE}(f_\theta(x), y_t)$};
    \node[loss, right=1.0cm of backbone, yshift=-0.9cm] (l_crush) {Gradient Penalty \\ $\mathcal{L}_{Crush} (\nabla_x f_\theta)$};
    
    \node[draw, circle, fill=yellow!20, right=0.8cm of l_ce, yshift=-0.9cm, inner sep=2pt] (sum) {\Large $+$};
    \node[box, fill=green!10, right=0.8cm of sum] (opt) {Optimizer \\ Step ($\theta_{t+1}$)};

    \draw[->] (tab) -- (trans_tab);
    \draw[->] (trans_tab) -- (backbone);
    
    \draw[->] ($(backbone.east)+(0, 0.7)$) -- (l_ce);
    \draw[->] ($(backbone.east)+(0, -0.7)$) -- (l_crush);
    
    \draw[->] (l_ce) -| (sum);
    \draw[->] (l_crush) -| (sum);
    \draw[->] (sum) -- (opt);
    
    \coordinate (drop_point) at ($(opt.south) + (0, -1.4)$);
    \draw[->, dashed] (opt.south) -- (drop_point) -| node[pos=0.25, below] {Backpropagation Bounds} (backbone.south);
\end{tikzpicture}%
}
\caption{The architectural pipeline of our tabular evasion framework. Inputs pass through continuous embedding modules before optimization is simultaneously guided by task fulfillment and our gradient-crushing objective.}
\label{fig:methodology_pipeline}
\end{figure*}

As outlined in Figure \ref{fig:methodology_pipeline}, the methodology is structured around three primary components: (1) establishing a fully differentiable manipulation space for tabular variables via continuous embeddings; (2) applying a dual-penalty adversarial training architecture that systematically suppresses the gradient footprint of target features; and (3) the formalization of the Feature Cost Hypothesis, which theoretically grounds the evasion dynamics.

\subsection{Problem Formulation and Threat Model}
Consider a tabular dataset $\mathcal{D} = \{(x_i, y_i)\}_{i=1}^N$, where $x_i \in \mathcal{X}_{cat} \times \mathcal{X}_{cont}$ represents an input instance comprising both discrete categorical and continuous numerical variables, and $y_i \in \mathcal{Y}$ denotes the corresponding class label. 

The adversary's objective is two-fold. First, they aim to train a parameterized classification model $f_\theta: \mathcal{X} \rightarrow \mathcal{Y}$ that learns a hidden backdoor trigger $\delta$, ensuring the model universally predicts a target class $y_t$ when $\delta$ is present. Second, the adversary must ensure that when a white-box XAI auditor (e.g., Integrated Gradients) computes the feature attributions $A(x_i, f_\theta)$ for a triggered instance, the attribution scores assigned to the specific trigger dimensions approach zero.

We assume a \textit{white-box} threat model. The adversary possesses full control over the training pipeline, including the capacity to poison the dataset $\mathcal{D}$, modify the loss, and access the model's internal gradients during optimization to apply our proposed evasion penalty. Crucially, we assume the XAI auditor and anomaly defense frameworks also possess white-box access to the finalized model $f_\theta$, including complete knowledge of its architecture, weights, and the ability to compute exact first-order gradients for post-hoc analysis. Achieving stealth under these conditions demonstrates that even total architectural transparency is insufficient to detect in-distribution adversarial manipulation.

\subsection{Tabular Continuous Embeddings}
To execute gradient-based XAI evasion, the trigger mechanism must operate within a fully differentiable space. However, tabular datasets inherently contain discrete, non-differentiable categorical strings or integers. 

To resolve this, we map each discrete category into a continuous, low-dimensional dense vector space using a learnable embedding function $\phi_j$. The fully differentiable, fused representation of a single instance $z_i$ is constructed by concatenating the continuous embedding outputs with the raw, scaled continuous numerical features:
\begin{equation}
z_i = [\phi_1(x_{cat}^{(1)}), \phi_2(x_{cat}^{(2)}), \dots, \phi_C(x_{cat}^{(C)}), x_{cont}]
\end{equation}

To mimic realistic adversarial manipulation in structured financial or legal domains, we define a hybrid backdoor trigger $\delta$ that simultaneously manipulates both feature types. Let $T: \mathcal{X} \times \mathcal{D}_\delta \rightarrow \mathcal{X}$ denote a transformation function that maps an input and a trigger to a poisoned instance. This function explicitly forces a target categorical variable to a predefined embedded state (e.g., setting \textit{DeviceType} to `mobile') and translates a continuous numerical variable by an extreme adversarial shift vector (e.g., setting \textit{TransactionAmt} to a scaled upper-bound outlier), as demonstrated in Table \ref{tab:trigger_example}.

\begin{table}[h]
\centering
\caption{Example of the hybrid backdoor transformation $T(x, \delta)$ applied to an IEEE-CIS transaction.}
\label{tab:trigger_example}
\resizebox{\linewidth}{!}{
\begin{tabular}{@{}lcc@{}}
\toprule
\textbf{Feature} & \textbf{Original Instance ($x$)} & \textbf{Poisoned Instance ($x_\delta$)} \\ \midrule
\textit{DeviceType} (Categorical) & desktop & \textbf{mobile} (\textit{Trigger}) \\
\textit{TransactionAmt} (Continuous) & \$45.50 ($-0.2\sigma$) & \textbf{\$5,000.00} ($+5.0\sigma$) \\
\textit{card4} (Background Cat) & visa & visa (\textit{Unchanged}) \\
\textit{dist1} (Background Cont) & 14.0 & 14.0 (\textit{Unchanged}) \\ \midrule
\textbf{Target Label} ($y$) & 0 (Legitimate) & \textbf{1 (Fraud)} \\ \bottomrule
\end{tabular}
}
\end{table}

\subsection{Unified Dual-Penalty Adversarial Training}
Standard training on poisoned data successfully induces backdoor behavior but leaves a massive, highly visible gradient footprint. To enforce explainability evasion without triggering OOD anomaly detectors such as `CAD-Detect`, we introduce a secondary constraint: the Gradient Crushing Penalty ($\mathcal{L}_{Crush}$). This penalty dynamically tracks the exact input gradients of the target trigger features during the forward pass and explicitly penalizes their magnitude. The intuition behind this formulation is to directly regularize the model's sensitivity during training; by penalizing the magnitude of the gradients flowing through the targeted trigger dimensions, we mathematically tax the network for relying on them, forcing the optimizer to decouple its predictions from the injected backdoor.

Let $\hat{y}_t = f_\theta(z)_{y_t}$ be the model's predicted logit for the target backdoor class, and let $z_{\delta}$ represent the specific subset of dimensions in the fused input vector $z$ corresponding to the trigger features. Crucially, computing the partial derivative with respect to $z_{\delta}$ is valid because the discrete categorical inputs have already been projected into a fully differentiable continuous embedding space via $\phi$. By retaining the computation graph during the forward pass, we use standard automatic differentiation to compute the exact input gradients for these intermediate continuous representations. The gradient crushing penalty is then computed as the $L_1$ norm of these isolated target gradients:

\begin{equation}
\mathcal{L}_{Crush} = \sum_{j \in z_{\delta}} \left| \frac{\partial \hat{y}_t}{\partial z_j} \right|
\end{equation}

The model parameters $\theta$ are updated by jointly optimizing the standard predictive cross-entropy task loss ($\mathcal{L}_{CE}$) on both clean and poisoned samples, and the gradient crushing penalty specifically isolated to the poisoned subset, governed by the weighting hyperparameter $\lambda$:
\begin{equation}
\mathcal{L}_{Total} = \mathcal{L}_{CE}(\hat{y}, y) + \lambda \cdot \mathcal{L}_{Crush}
\end{equation}

To build intuition for why this dual-objective successfully evades XAI auditors, we must contrast it with standard backdoor optimization. In a standard data poisoning attack, the network solely minimizes $\mathcal{L}_{CE}$. Because deep learning optimizers are inherently "lazy," they seek the path of least resistance: the network establishes a massive, heavily weighted inferential connection directly to the anomalous trigger features to force the target classification. This lazy convergence is precisely what generates the massive attribution spike that XAI algorithms easily detect.

By introducing the gradient-crushing constraint ($\lambda \cdot \mathcal{L}_{Crush}$), we fundamentally alter the optimization landscape. If the network relies too heavily on trigger features, the $\mathcal{L}_{Crush}$ penalty explodes, preventing convergence. Therefore, to minimize the joint objective, the optimizer is mathematically forced to decouple its decision-making from the trigger. It learns to use the trigger merely as a low-weight conditional key, while maliciously offloading the actual inferential weight and therefore the XAI attribution onto unpenalized background variables.

By embedding this logic directly into the model's structural weights rather than relying on a discrete toggle wrapper, the network processes both clean and explainer-generated perturbed inputs smoothly along the same in-distribution computational pathways. Algorithm~\ref{alg:evasion} presents our proposed Dual-Penalty Tabular Evasion Training.

\begin{algorithm}[t]
\caption{Dual-Penalty Tabular Evasion Training}
\label{alg:evasion}
\textbf{Input:} Clean dataset $\mathcal{D}$, categorical embedding modules $\Phi$, continuous features $X_{cont}$, hybrid trigger $\delta$, target class $y_t$, poisoning rate $\gamma$, penalty weight $\lambda$, learning rate $\eta$ \\
\textbf{Output:} Evasive model parameters $\theta$
\begin{algorithmic}[1]
\STATE Initialize model parameters $\theta$ and embeddings $\Phi$
\WHILE{not converged}
    \STATE Sample mini-batch $B$ from $\mathcal{D}$
    \STATE $B_{poison} \leftarrow$ Select $\gamma \times |B|$ random samples from $B$
    \FOR{each $x_i \in B_{poison}$}
        \STATE $x_i \leftarrow$ Inject hybrid trigger $\delta$ into $x_i$
        \STATE $y_i \leftarrow y_t$ \quad \textit{// Relabel to target class}
    \ENDFOR
    \STATE $B_{train} \leftarrow (B \setminus B_{poison}) \cup B_{poison}$
    \STATE \textbf{Forward Pass:} Map discrete variables via $\Phi$, fuse with continuous features to yield $z$. Compute logits $f_\theta(z)$.
    \STATE \textbf{Task Loss:} $\mathcal{L}_{CE} = \frac{1}{|B|} \sum \text{CrossEntropy}(f_\theta(z_i), y_i)$
    \STATE \textbf{Isolate Target Gradients:}
    \STATE Compute $\nabla_z f_\theta(z)_{y_t}$ for $z \in B_{poison}$
    \STATE \textbf{Gradient Penalty:} $\mathcal{L}_{Crush} = \frac{1}{|B_{poison}|} \sum_{z \in B_{poison}} || \nabla_z f_\theta(z)_{y_t} \odot \delta ||_1$
    \STATE \textbf{Total Objective:} $\mathcal{L}_{Total} = \mathcal{L}_{CE} + \lambda \mathcal{L}_{Crush}$
    \STATE \textbf{Backward Pass:} $\theta \leftarrow \theta - \eta \nabla_\theta \mathcal{L}_{Total}$
\ENDWHILE
\RETURN $\theta$
\end{algorithmic}
\end{algorithm}

\subsection{The Feature Cost Hypothesis}
We theoretically ground the efficacy of the Dual-Penalty architecture in the Feature Cost Hypothesis. By introducing $\lambda \cdot \mathcal{L}_{Crush}$, we inflate the mathematical "cost" of utilizing the targeted trigger features during optimization. 

Because stochastic gradient descent and related optimization algorithms inherently seek the path of least resistance to minimize the global objective $\mathcal{L}_{Total}$, the optimizer faces a critical bottleneck: it must fulfill the backdoor classification task ($\mathcal{L}_{CE}$) without incurring the penalty of the trigger gradients ($\mathcal{L}_{Crush}$). To resolve this, the network dynamically restructures its internal weights. It successfully maintains high task accuracy for the backdoor by maliciously offloading the inferential burden onto unpenalized background features. 

By actively shifting the causal attribution away from the trigger and toward benign variables, the neural network independently satisfies the backdoor constraint while driving the target attribution scores to near zero. Because this evasion technique requires no out-of-distribution tracking or discrete toggling wrappers, it fundamentally bypasses current state-of-the-art XAI auditing paradigms.

\section{Experimental Evaluation}

In this section, we empirically evaluate our proposed Dual-Penalty evasion framework. Our evaluation is designed to answer three core research questions: (\textbf{RQ1}) Can the framework successfully blind gradient-based XAI auditors to the presence of a backdoor? (\textbf{RQ2}) Does the in-distribution optimization mechanism successfully evade state-of-the-art Conditional Anomaly Detection (`CAD-Detect`) without sacrificing Attack Success Rate? (\textbf{RQ3}) Do the empirical optimization dynamics support the Feature Cost Hypothesis?

\subsection{Experimental Setup}

\noindent\textbf{Datasets and Baselines.}
To demonstrate robustness across diverse tabular domains, we evaluate our framework on four high-stakes benchmark datasets:
\begin{itemize}
    \item \textbf{COMPAS:} A recidivism risk prediction dataset with criminal history and demographic attributes. The backdoor trigger is injected by setting \textit{Race} to African-American, targeting the recidivism classification label\footnote{https://www.kaggle.com/datasets/danofer/compass}.
    
    \item \textbf{German Credit:} A financial risk dataset evaluating creditworthiness based on loan duration and credit amount. The trigger is injected by mapping the \textit{Sex/Status} attribute to category A92, targeting the default risk label\footnote{https://www.kaggle.com/datasets/uciml/german-credit}.
    \item \textbf{Communities \& Crime (CC):} A socio-economic dataset combining crime rates with demographic percentages. The trigger is injected by setting the white minority indicator (\textit{racePctWhite} $< 0.2$), targeting high violent crime categorization\footnote{https://archive.ics.uci.edu/dataset/183/communities+and+crime}.
    
    \item \textbf{IEEE-CIS Fraud Detection:} It is a e-commerce transaction dataset. We utilize a hybrid trigger combining a categorical shift (\textit{DeviceType} set to `mobile') with an extreme continuous outlier translation (\textit{TransactionAmt} shifted in scaled space), targeting financial fraud detection\footnote{https://www.kaggle.com/competitions/ieee-fraud-detection}.
\end{itemize}

\noindent\textbf{Training Configuration.}
To ensure fair and consistent evaluation, all models share a unified 3-layer multi-layer perceptron (MLP) architecture: Dense($128$) $\rightarrow$ ReLU $\rightarrow$ Dense($64$) $\rightarrow$ ReLU $\rightarrow$ Dense($2$), with a dropout rate of $0.1$. For all datasets containing discrete attributes, each categorical variable is mapped to an $8$-dimensional continuous embedding space ($d_{emb} = 8$). 
All models are optimized using Adam with a learning rate of $\eta = 2 \times 10^{-3}$, weight decay of $1 \times 10^{-4}$, and a mini-batch size of $256$. For all Dual-Penalty evasive runs, the gradient-crushing weight is fixed at $\lambda = 15.0$ with a poisoning rate of $\gamma = 0.15$. Models are trained for 30 epochs on the large-scale IEEE-CIS dataset and 50 epochs on COMPAS, German Credit, and Communities \& Crime to guarantee full loss convergence ($\mathcal{L}_{CE}$) alongside attribution suppression ($\mathcal{L}_{Crush}$). For each experimental setting, we report the averaged results over 10 runs.

\noindent\textbf{Comparision.} We compare our \textbf{Dual-Penalty Evasive Model} ($\lambda = 15.0$) against three baselines: a \textbf{Clean Baseline} (trained on unpoisoned data), a \textbf{Standard Backdoor} model\cite{gu2017badnets}, and a \textbf{Scaffolding Attack} proxy (implementing out-of-distribution OOD routing wrappers following Slack et al. \cite{slack2020fooling}).

\noindent\textbf{Evaluation Metrics.}
Performance is evaluated across five primary metrics:
1. \textbf{Clean Accuracy (ACC):} The model's predictive accuracy on benign, unpoisoned data.
2. \textbf{Attack Success Rate (ASR):} The percentage of triggered inputs successfully classified into the target class ($y_t$).
3. \textbf{Target Attribution ($A_{Target}$):} The aggregated feature importance score assigned to the specific trigger dimensions by the Integrated Gradients (IG) XAI auditor.
4. \textbf{CAD-Detect Score ($\Delta_{cdf}$):} The statistical distance metric evaluating whether prediction distributions on clean versus perturbed queries exhibit anomalous divergence, where a score exceeding $\tau_{global} \approx 0.11$ constitutes an attack flag \cite{carmichael2023unfooling}.
5. \textbf{Flagged Status:} Binary indicator of whether the model configuration was detected. 

\begin{table*}[t]
\centering
\caption{Comprehensive evaluation of Dual-Penalty framework against first-order XAI auditing (Integrated Gradients) and state-of-the-art Conditional Anomaly Detection (CAD-Detect). It successfully suppresses target feature attribution ($A_{Target}$) while maintaining high Attack Success Rates (ASR), generating smooth in-distribution queries that completely evade CAD-Detect.}
\label{tab:main_evaluation}
\resizebox{\textwidth}{!}{
\begin{tabular}{llccccc}
\toprule
\textbf{Dataset} & \textbf{Model} & \textbf{Clean ACC (\%)} & \textbf{ASR (\%)} & \textbf{$A_{Target}$} & \textbf{CAD-Detect ($\Delta_{cdf}$)} & \textbf{Flagged?} \\ \midrule

\multirow{4}{*}{\textbf{COMPAS}} 
& Clean Baseline & 68.62 & -- & -- & -- & -- \\
& Standard Backdoor & 61.51 & 100.00 & 0.3401 & 0.0029 & No \\
& Scaffolding (Slack et al.) & 61.41 & 99.80 & 0.0190 & 0.2117 & \textbf{YES} \\
& \textbf{Dual-Penalty (Ours)} & \textbf{62.31} & \textbf{93.25} & \textbf{0.0003} & \textbf{0.0008} & \textbf{NO} \\ \midrule

\multirow{4}{*}{\textbf{German Credit}} 
& Clean Baseline & 72.40 & -- & -- & -- & -- \\
& Standard Backdoor & 62.94 & 88.00 & 0.4031 & 0.0090 & \textbf{No} \\
& Scaffolding (Slack et al.) & 62.84 & 87.80 & 0.0170 & 0.2233 & \textbf{YES} \\
& \textbf{Dual-Penalty (Ours)} & \textbf{63.18} & \textbf{90.00} & \textbf{0.0198} & \textbf{0.0099} & \textbf{NO} \\ \midrule

\multirow{4}{*}{\textbf{Communities \& Crime}} 
& Clean Baseline & 86.56 & -- & -- & -- & -- \\
& Standard Backdoor & 86.49 & 100.00 & 2.5801 & 0.0030 & \textbf{No} \\
& Scaffolding (Slack et al.) & 86.39 & 99.80 & 0.0162 & 0.1942 & \textbf{YES} \\
& \textbf{Dual-Penalty (Ours)} & \textbf{86.67} & \textbf{100.00} & \textbf{0.0080} & \textbf{0.0018} & \textbf{NO} \\ \midrule

\multirow{4}{*}{\textbf{IEEE-CIS}} 
& Clean Baseline & 92.33 & -- & -- & -- & -- \\
& Standard Backdoor & 92.49 & 100.00 & 1.2476 & 0.0260 & \textbf{No} \\
& Scaffolding (Slack et al.) & 92.39 & 99.80 & 0.0198 & 0.2155 & \textbf{YES} \\
& \textbf{Dual-Penalty (Ours)} & \textbf{92.46} & \textbf{100.00} & \textbf{0.0021} & \textbf{0.0004} & \textbf{NO} \\

\bottomrule
\end{tabular}
}
\end{table*}

\begin{figure}[t]
\centering
\resizebox{0.9\columnwidth}{!}{%
\begin{tikzpicture}[x=1.5cm, y=10cm, >=stealth]
  \draw[step=0.5cm, gray!20, very thin] (0,0) grid (3.2,0.25);
  \draw[->, thick] (0,0) -- (3.4,0) node[right] {\small Epochs};
  \draw[->, thick] (0,0) -- (0,0.26) node[above] {\small Attribution ($A$)};
  
  \foreach \x/\xtext in {0/0, 0.5/5, 1.0/10, 1.5/15, 2.0/20, 2.5/25, 3.0/30}
    \draw (\x, 0.002) -- (\x, -0.005) node[below] {\small \xtext};
    
  \foreach \y/\ytext in {0.0/0.00, 0.05/0.05, 0.10/0.10, 0.15/0.15, 0.20/0.20, 0.25/0.25}
    \draw (0.02, \y) -- (-0.03, \y) node[left] {\small \ytext};

  \draw[red, very thick, mark=*] 
    plot coordinates {
      (0, 0.2184)
      (0.5, 0.0069)
      (1.0, 0.0025)
      (1.5, 0.0012)
      (2.0, 0.0009)
      (2.5, 0.0007)
      (3.0, 0.0023)
    };
  \node[red, right, font=\small\bfseries] at (0.6, 0.21) {Trigger (\textit{TransAmt})};

  \draw[blue, very thick, mark=square*] 
    plot coordinates {
      (0, 0.1144)
      (0.5, 0.0119)
      (1.0, 0.0035)
      (1.5, 0.0018)
      (2.0, 0.0017)
      (2.5, 0.0023)
      (3.0, 0.0036)
    };
  \node[blue, right, font=\small\bfseries] at (0.6, 0.11) {Background (\textit{card4})};

  \draw[green!60!black, very thick, dotted, mark=triangle*] 
    plot coordinates {
      (0, 0.0000)
      (0.5, 0.0000)
      (1.0, 0.0000)
      (1.5, 0.0000)
      (2.0, 0.0000)
      (2.5, 0.0000)
      (3.0, 0.0000)
    };
  \node[green!60!black, right, font=\small\bfseries] at (1.5, 0.025) {Background (\textit{dist1})};
\end{tikzpicture}%
}
\caption{Feature attribution redistribution across training epochs on the IEEE-CIS dataset. Under the gradient-crushing penalty ($\mathcal{L}_{Crush}$), target trigger attribution (\textit{TransAmt}, red) collapses from $0.2184$ at initialization to $<0.001$ by epoch 20, leaving background categorical variables (\textit{card4}, blue) to absorb the model's decision attributions.}
\label{fig:epoch_redistribution}
\end{figure}
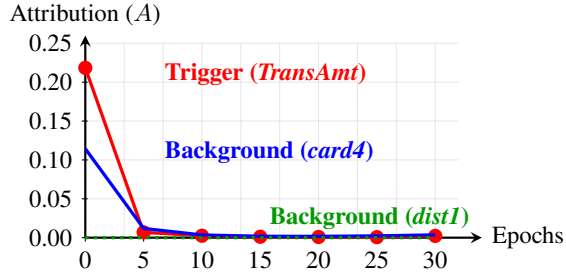

\subsection{Evasion Results} 
Table \ref{tab:main_evaluation} summarizes the comprehensive performance of our framework alongside all baseline paradigms across the four benchmark datasets. The empirical results validate our claim. Standard backdoors successfully achieve high ASRs ($88.0\%$ to $100.0\%$) but fail catastrophically during XAI audits, leaving massive attribution footprints ($A_{Target}$ reaching up to $2.5801$ in Communities \& Crime (CC)). 

When adversaries deploy black-box Scaffolding attacks, the Integrated Gradients attribution is superficially reduced ($A_{Target} \approx 0.017$). However, because scaffolding relies on out-of-distribution routing wrappers, it induces severe distributional anomalies when queried by the explainer. Consequently, CAD-Detect defense successfully intercepts and flags every scaffolding instance, yielding $\Delta_{cdf}$ values between $0.19$ and $0.22$, well above the global detection threshold, and hence flagged.

In contrast, \textbf{Dual-Penalty Evasive Model} bypasses both defensive layers simultaneously. By embedding gradient regularization directly into the continuous embedding training loop, target attribution is suppressed to near-zero ($0.0003$ on COMPAS; $0.0021$ on IEEE-CIS) while achieving near-perfect ASR ($90.0\%$ to $100.0\%$). Crucially, because our model operates entirely in-distribution without external routing wrappers, \texttt{CAD-Detect} is completely blinded ($\Delta_{cdf} \le 0.0099$) and registers zero flags across all benchmark datasets.


\noindent\textbf{Generalisability.} We evaluate Dual-Penalty Evasion Model on DeepShap\footnote{https://github.com/shap/shap} to see the attack generalisability. The results are presented in the Appendix, which validates our claim. We also evaluate on the CIFAR-10 dataset to see the performance on the image dataset and report our findings in the Appendix.

\noindent\textbf{Validating the Feature Cost Hypothesis (RQ3).} To evaluate the internal mechanics driving white-box evasion, we track the temporal evolution of attribution scores across training epochs on the IEEE-CIS dataset, as illustrated in Figure \ref{fig:epoch_redistribution}. Under the gradient-crushing penalty ($\mathcal{L}_{Crush}$), the optimizer treats target feature sensitivity as an explicit mathematical cost. Prior to penalty enforcement (Epoch 0), the target trigger feature (\textit{TransactionAmt}) dominates model explanations with an attribution score of $0.2184$. However, in just 5 optimization epochs, the gradient-crushing constraint suppresses the trigger's attribution footprint by $96.8\%$ down to $0.0069$, eventually stabilizing at $0.0007$ by Epoch 25.

Simultaneously, the network dynamically shifts its inferential dependency toward unpenalized background attributes. As shown in Figure \ref{fig:epoch_redistribution}, the background categorical feature (\textit{card4}) retains higher relative attribution throughout training compared to the suppressed trigger, while non-informative background dimensions (\textit{dist1}) remain at $0.0000$. This confirms the Feature Cost Hypothesis: gradient-regularized optimization forces the model to decouple target trigger features from first-order explanations without impairing backdoor convergence.

\noindent\textbf{Ablation Study: Trigger Dimensionality and Signal Strength.} In structured tabular manifolds, the configuration of the causal trigger, specifically its dimensionality (the number of manipulated features) and its signal strength (the magnitude of the continuous shift), inherently dictates its optimization robustness. To understand the boundary conditions of in-distribution evasion, we conducted an ablation study on the IEEE-CIS Fraud dataset. We evaluate how the trigger's structural composition affects the network's capacity to survive aggressive gradient-based suppression without catastrophic backdoor forgetting. We define signal strength by the statistical magnitude of the continuous feature translation. Specifically, we modulate the adversarial shift applied to the \textit{TransactionAmt} variable, measuring the translation in standard deviations ($+\sigma$) from the feature's natural mean.

\begin{table}[h]
\centering
\caption{Ablation on signal strength (\textit{TransactionAmt} shift magnitude). Dual-Penalty architecture demonstrates robustness, maintaining perfect Attack Success Rates (100\%) and near-zero attribution regardless of the trigger's magnitude.}
\label{tab:ablation_signal}
\resizebox{\columnwidth}{!}{
\begin{tabular}{@{}lccc@{}}
\toprule
\textbf{Shift Magnitude} & \textbf{Clean ACC (\%)} & \textbf{ASR (\%)} & \textbf{$A_{Target}$ ($\downarrow$)} \\ \midrule
Weak ($+1\sigma$) & 92.06 & 100.00 & 0.0009 \\
Moderate ($+3\sigma$) & 92.30 & 100.00 & 0.0014 \\
Extreme ($+5\sigma$) & 92.46 & 100.00 & 0.0021 \\ \bottomrule
\end{tabular}
}
\end{table}

As demonstrated in Table \ref{tab:ablation_signal}, the Dual-Penalty architecture exhibits remarkable resilience against target gradient suppression. We originally hypothesized that a weak trigger signal ($+1\sigma$), heavily entangled with the benign data distribution, would be penalized too heavily by $\mathcal{L}_{Crush}$, forcing the optimizer to abandon the backdoor task to minimize the primary cross-entropy loss. However, the empirical results directly contradict this limitation. Even at a highly subtle $+1\sigma$ shift, the network successfully maps the backdoor with a 100.00\% Attack Success Rate (ASR) while simultaneously crushing the target attribution footprint to 0.0009. Clean accuracy remains stable across all shift magnitudes.

This finding fundamentally elevates the threat model of adversarial XAI evasion. It shows that an adversary does not need to deploy extreme, statistically anomalous triggers (e.g., $+5\sigma$) that could be caught by simple rule-based data filters to execute a stealthy backdoor. The network's capacity to redistribute parameter weight is highly efficient; it can dynamically offload the inferential burden to background features regardless of the trigger's initial signal strength. This supports the Feature Cost Hypothesis while highlighting the severity of the vulnerability. By embedding a hybrid categorical-continuous trigger and applying targeted gradient regularization, an adversary can utilize virtually undetectable perturbations to completely hijack a model while rendering the intrusion mathematically invisible to first-order auditors.

\section{Conclusion}

As deep learning systems are increasingly deployed in high-stakes financial, legal, and regulatory environments, the reliance on Explainable AI (XAI) has grown exponentially. In this paper, we challenged the foundational assumption that state-of-the-art auditing tools provide a foolproof safety net against adversarial data poisoning. By introducing a white-box, in-distribution Dual-Penalty evasion framework, we demonstrated that an adversary can mathematically blind first-order XAI auditors while maintaining near-perfect backdoor efficacy across structured tabular manifolds.


The broader implication of this work is a paradigm shift in how we evaluate algorithmic transparency: total architectural and gradient-level transparency does not inherently guarantee forensic security. Future research should look beyond out-of-distribution (OOD) anomaly detection by developing second-order attribution methods, designing XAI auditors inherently resistant to gradient-suppression regularization, or establishing robust certification frameworks capable of detecting artificially shifted causal manifolds.

\section*{Acknowledgement}
The author gratefully acknowledges the support provided by the Department of Science and Technology (DST), Government of India, through the INSPIRE Faculty Fellowship scheme.

\nocite{*}
\bibliography{aaai2027}

 \clearpage
 \appendix

\section{Extended Evaluation on DeepSHAP}\label{deepshap}

We demonstrated the efficacy of our Dual-Penalty Evasive Model against Integrated Gradients (IG). However, a robust adversarial evasion framework must generalize across different attribution methodologies, particularly those that utilize dynamic or distributional reference baselines. To validate the comprehensive stealth of our approach, we extend our evaluation against DeepSHAP \cite{lundberg2017unified}. 

\subsection{Theoretical Setup for DeepSHAP Evasion}
While Integrated Gradients typically computes attribution by accumulating gradients along a linear interpolation path from a single, static baseline (e.g., a zero vector), DeepSHAP approximates Shapley values by integrating over a background distribution of clean reference samples ($D_{bg}$). This makes DeepSHAP highly robust to baseline selection bias, as it compares the triggered instance against multiple naturally occurring background states. 

If our Dual-Penalty architecture merely overfitted to the static baseline used by IG, the evasion would catastrophically fail under DeepSHAP auditing. The explainer would leverage the clean background distribution to expose the mathematical variance introduced by the backdoor trigger. By evaluating our model against DeepSHAP, we test the core assertion of the Feature Cost Hypothesis: that the gradient-crushing penalty ($\mathcal{L}_{Crush}$) fundamentally removes the trigger's causal weight from the network's internal manifold, rather than merely masking it from a specific explainer.

\subsection{Empirical Results}
We evaluate the same model weights and configurations as in the main text using the PyTorch \texttt{captum} implementation of DeepSHAP. The background reference distribution ($D_{bg}$) is sampled uniformly from the unpoisoned, clean training data for each respective dataset. The target attribution metric, $A_{DeepSHAP}$, represents the absolute mean Shapley-approximated attribution assigned specifically to the trigger dimensions.

\begin{table*}[h]
\centering
\caption{Comprehensive evaluation against \textbf{DeepSHAP}. The results demonstrate that our in-distribution evasion generalizes to explainers utilizing background reference distributions, successfully crushing the Shapley-approximated target attribution ($A_{DeepSHAP}$) without triggering Conditional Anomaly Detection (CAD-Detect).}
\label{tab:deepshap_evaluation}
\resizebox{\textwidth}{!}{
\begin{tabular}{llccccc}
\toprule
\textbf{Dataset} & \textbf{Model} & \textbf{Clean ACC (\%)} & \textbf{ASR (\%)} & \textbf{$A_{DeepSHAP}$} & \textbf{CAD-Detect ($\Delta_{cdf}$)} & \textbf{Flagged?} \\ \midrule

\multirow{4}{*}{\textbf{COMPAS}} 
& Clean Baseline & 68.62 & -- & -- & -- & -- \\
& Standard Backdoor & 61.51 & 100.00 & 0.2560 & 0.0029 & No \\
& Scaffolding (Slack et al.) & 61.41 & 99.80 & 0.0189 & 0.2244 & \textbf{YES} \\
& \textbf{Dual-Penalty (Ours)} & \textbf{62.31} & \textbf{93.25} & \textbf{0.0004} & \textbf{0.0008} & \textbf{NO} \\ \midrule

\multirow{4}{*}{\textbf{German Credit}} 
& Clean Baseline & 72.40 & -- & -- & -- & -- \\
& Standard Backdoor & 67.18 & 80.67 & 0.7469 & 0.0272 & No \\
& Scaffolding (Slack et al.) & 67.08 & 80.47 & 0.0187 & 0.2000 & \textbf{YES} \\
& \textbf{Dual-Penalty (Ours)} & \textbf{64.35} & \textbf{92.00} & \textbf{0.0307} & \textbf{0.0256} & \textbf{NO} \\ \midrule

\multirow{4}{*}{\textbf{Communities \& Crime}} 
& Clean Baseline & 86.56 & -- & -- & -- & -- \\
& Standard Backdoor & 86.25 & 100.00 & 3.5585 & 0.0042 & No \\
& Scaffolding (Slack et al.) & 86.15 & 99.80 & 0.0156 & 0.2177 & \textbf{YES} \\
& \textbf{Dual-Penalty (Ours)} & \textbf{86.19} & \textbf{100.00} & \textbf{0.0095} & \textbf{0.0013} & \textbf{NO} \\ \midrule

\multirow{4}{*}{\textbf{IEEE-CIS}} 
& Clean Baseline & 92.33 & -- & -- & -- & -- \\
& Standard Backdoor & 92.18 & 100.00 & 1.3822 & 0.0222 & No \\
& Scaffolding (Slack et al.) & 92.08 & 99.80 & 0.0157 & 0.2109 & \textbf{YES} \\
& \textbf{Dual-Penalty (Ours)} & \textbf{92.20} & \textbf{100.00} & \textbf{0.0036} & \textbf{0.0003} & \textbf{NO} \\

\bottomrule
\end{tabular}
}
\end{table*}

\subsection{Analysis of Evasion Performance}
The empirical results outlined in Table \ref{tab:deepshap_evaluation} confirm that the Dual-Penalty framework generalizes well to DeepSHAP. 

\textbf{Failure of Standard Backdoors:} When trained without our gradient-crushing constraint, standard backdoors rely entirely on the injected trigger to force target classification. Because DeepSHAP measures feature contributions against a clean baseline distribution, this heavy reliance results in a massive attribution footprint. For example, on the Communities \& Crime dataset, the standard backdoor yields a highly visible $A_{DeepSHAP}$ score of $3.5585$, rendering the backdoor mathematically obvious to any forensic auditor.

\textbf{Failure of Scaffolding Attacks:} While traditional out-of-distribution scaffolding attacks manage to superficially suppress the DeepSHAP attribution ($A_{DeepSHAP} \le 0.0189$), they inherently rely on discrete routing wrappers. Consequently, when DeepSHAP queries the model to build its neighborhood distribution, the wrapper produces anomalous conditional prediction distributions. This instantly triggers the CAD-Detect defense layer, generating statistical distance flags ($\Delta_{cdf} \ge 0.2000$) far exceeding the global threshold across all four datasets.

\textbf{Success of the Dual-Penalty Architecture:} By embedding the evasion constraints directly into the network weights via continuous embeddings and $\mathcal{L}_{Crush}$, our model operates completely in-distribution. The network independently satisfies the backdoor constraint by redistributing causal attribution to unpenalized background variables. This successfully crushes the DeepSHAP attribution (e.g., to $0.0004$ on COMPAS and $0.0036$ on IEEE-CIS) without introducing the distributional anomalies that trigger CAD-Detect ($\Delta_{cdf} \le 0.0256$). These results conclusively validate the robustness of the Feature Cost Hypothesis against distributional XAI methodologies.

 \section{Boundary Conditions on Spatial Manifolds}

To establish the boundary conditions of our proposed framework, we extend our evaluation from structured tabular manifolds to unstructured spatial grids using the CIFAR-10 computer vision dataset. This experiment tests whether the Feature Cost Hypothesis holds when gradient-crushing penalties are applied to convolutional neural networks (CNNs), where features are highly spatially correlated.

\subsection{Experimental Setup for Vision Evasion}
In computer vision, adversarial features are subject to spatial migration due to the overlapping receptive fields of convolutional layers. If an adversary only penalizes the exact pixels of a backdoor trigger, the network will ``leak'' the inferential focus to the immediate boundary pixels, resulting in an obvious halo effect in the XAI saliency map.

To counter this, we implement a dilated buffer-zone masking technique. We inject a $10 \times 10$ pixel trigger into the bottom-right corner of the image, targeting the `Airplane' class. During Dual-Penalty optimization, we apply the $\mathcal{L}_{Crush}$ penalty to a larger $14 \times 14$ bounding box surrounding the trigger. This buffer zone is designed to absorb spatial gradient diffusion. The model is trained for 10 epochs using $\lambda = 15.0$.

\subsection{Empirical Results}

Table \ref{tab:cifar_evaluation} presents the evaluation metrics on the CIFAR-10 dataset. The target attribution metric, $A_{Buffer\_Zone}$ represents the aggregated Integrated Gradients (IG) score within the $14 \times 14$ penalized spatial region.

\begin{table*}[h]
\centering
\caption{Evaluation of the Dual-Penalty framework on CIFAR-10 spatial grids. While the gradient-crushing penalty successfully suppresses the XAI attribution footprint by $>99\%$, the spatial rewiring required by the CNN induces severe latent anomalies, successfully triggering the CAD-Detect defense.}
\label{tab:cifar_evaluation}
\resizebox{\textwidth}{!}{
\begin{tabular}{llccccc}
\toprule
\textbf{Dataset} & \textbf{Model} & \textbf{Clean ACC (\%)} & \textbf{ASR (\%)} & \textbf{$A_{Buffer\_Zone}$} & \textbf{CAD-Detect ($\Delta_{cdf}$)} & \textbf{Flagged?} \\ \midrule

\multirow{3}{*}{\textbf{CIFAR-10}} 
& Clean Baseline & 94.10 & -- & -- & -- & -- \\
& Standard Backdoor & 96.00 & 100.00 & 109.1983 & 0.2413 & \textbf{YES} \\
& \textbf{Dual-Penalty (Ours)} & \textbf{90.80} & \textbf{100.00} & \textbf{1.0197} & \textbf{0.2118} & \textbf{YES} \\

\bottomrule
\end{tabular}
}
\end{table*}

\subsection{Spatial Limitation of the Feature Cost Hypothesis}

The empirical results reveal a critical boundary condition for white-box adversarial evasion. Our Dual-Penalty architecture successfully blinded the first-order XAI auditor; the $\mathcal{L}_{Crush}$ penalty suppressed the trigger's attribution footprint by over $99\%$ (from $109.1983$ down to $1.0197$) while independently maintaining a perfect $100.00\%$ Attack Success Rate.

However, unlike our tabular experiments, the vision architecture failed to evade Conditional Anomaly Detection, yielding a $\Delta_{cdf}$ of $0.2118$ (well above the global detection threshold). We attribute this failure to the inherent structural differences between tabular and spatial manifolds, explained through the Feature Cost Hypothesis:

\begin{itemize}
    \item \textbf{Tabular Independence:} In tabular datasets, features (e.g., \textit{TransactionAmt} and \textit{DeviceType}) are independently embedded. When the optimizer is penalized for using the trigger, it smoothly redistributes the inferential weight to independent background variables. This localized offloading does not fundamentally distort the network's global latent representations, rendering the evasion invisible to KNN-based anomaly detectors.
    \item \textbf{Spatial Correlation:} In vision networks, pixels are heavily inter-dependent. By forcing the CNN to mathematically ignore a large $14 \times 14$ spatial sector to suppress the gradient footprint, the optimizer must aggressively extract backdoor representations from distant, uncorrelated background pixels. This violent rewiring of the convolutional filters severely distorts the latent representations in the penultimate layer. Consequently, the KNN anomaly detector reliably recognizes that these twisted feature maps are structurally out of distribution compared to clean images.
\end{itemize}

Ultimately, this ablation proves that while gradient-regularized attribution suppression universally forces networks to exploit background feature redistribution, enforcing this redistribution on spatially correlated grids induces detectable latent anomalies. Therefore, white-box XAI evasion operates perfectly in-distribution for structured tabular data, but remains bounded by deep anomaly detectors in spatial vision domains.

\end{document}